\documentclass[10pt,twocolumn,letterpaper]{article}

\usepackage{3dv}
\usepackage{times}
\usepackage{epsfig}
\usepackage{graphicx}
\usepackage{amsmath}
\usepackage{amssymb}

\usepackage[utf8]{inputenc} 
\usepackage[T1]{fontenc}    
\usepackage{url}            
\usepackage{booktabs}       
\usepackage{multirow}
\usepackage{multicol}
\usepackage{color}
\usepackage{wrapfig}
\usepackage{xcolor}
\usepackage{xspace}
\usepackage{caption}
\usepackage{tablefootnote}
\usepackage{epsfig}
\usepackage{makecell}
\usepackage{graphicx}

\usepackage{adjustbox}
\usepackage{array}
\usepackage{bbm}
\usepackage{tabularx}
\usepackage{subcaption}
\usepackage{animate}

\newcommand{\myparagraph}[1]{\vspace{0.0em}\noindent\textbf{#1}}
\usepackage[pagebackref=true,breaklinks=true,colorlinks,bookmarks=false]{hyperref}

\threedvfinalcopy 

\def\threedvPaperID{286} 

\ifthreedvfinal\fi
\begin{document}

\title{Direct Dense Pose Estimation}

\author{Liqian Ma$^{1}$ \quad Lingjie Liu$^{2}$ \quad Christian Theobalt$^{2}$ \quad Luc Van Gool$^{1,3}$ \\
\\
$^{1}$KU-Leuven/PSI, Toyota Motor Europe (TRACE) \\
$^{2}$Max Planck Institute for Informatics, Saarland Informatics Campus \quad  $^{3}$ETH Zurich \\
{\tt\small{\{liqian.ma,luc.vangool\}@esat.kuleuven.be}} \\
{\tt\small{\{theobalt, lliu\}@mpi-inf.mpg.de}} \quad  {\tt\small{vangool@vision.ee.ethz.ch}}
}

\maketitle
\thispagestyle{empty}

\begin{abstract}
Dense human pose estimation is the problem of learning dense correspondences between RGB images and the surfaces of human bodies, which finds various applications, such as human body reconstruction, human pose transfer, and human action recognition. Prior dense pose estimation methods are all based on Mask R-CNN framework and operate in a top-down manner of first attempting to identify a bounding box for each person and matching dense correspondences in each bounding box. Consequently, these methods lack robustness due to their critical dependence on the Mask R-CNN detection, and the runtime increases drastically as the number of persons in the image increases. We therefore propose a novel alternative method for solving the dense pose estimation problem, called {\em Direct Dense Pose (DDP)}. DDP first predicts the instance mask and global IUV representation separately and then combines them together. We also propose a simple yet effective 2D temporal-smoothing scheme to alleviate the temporal jitters when dealing with video data. Experiments demonstrate that DDP overcomes the limitations of previous top-down baseline methods and achieves competitive accuracy. In addition, DDP is computationally more efficient than previous dense pose estimation methods, and it reduces jitters when applied to a video sequence, which is a problem plaguing the previous methods.
\end{abstract}

\section{Introduction}
\label{sec:intro}

\begin{figure*}[t]
\scriptsize
  \centering
  \includegraphics[width=1\linewidth]{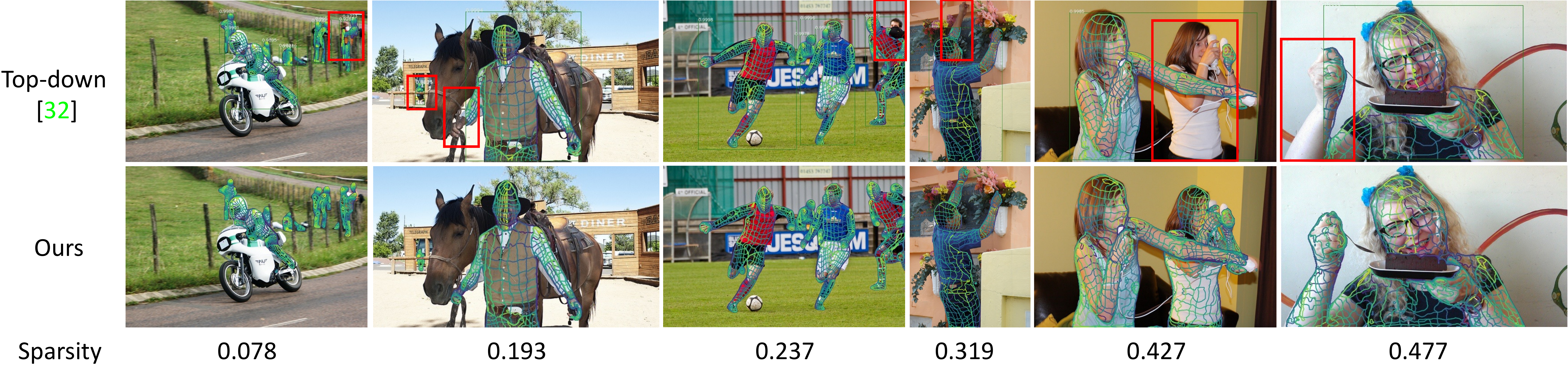}\\
\caption{Top: illustration of top-down methods~\cite{neverova2020continuous} early commitment and overlapping ambiguity issues (red boxes). Bottom: data sparsity, \ie the ratio of human pixels over the whole image pixels.
}
\label{fig:teaser}
\end{figure*}

Human dense pose estimation aims to learn the correspondence between RGB images and 3D human model surfaces.
It is a fundamental and essential problem in human-centric analysis and synthesis applications,
such as 3D body shape estimation~\cite{xu2019denserac,zhang2020learning}, human pose transfer~\cite{grigorev2019coordinate,sarkar2020neural}, unselfie~\cite{ma2020unselfie}, and character animation~\cite{gafni2019vid2game,wang2018vid2vid}.
G{\"u}ler~\etal ~\cite{alp2018densepose} 
densely map each person pixel of an `in the wild'' RGB image to a location of a 3D human model surface using a two-stage top-down framework based on Mask R-CNN~\cite{he2017mask}. 
They sparsely annotate a human subset of the COCO dataset~\cite{lin2014microsoft} for training. Follow-up works also adopt the same top-down principle to further improve the performance via uncertainty modeling~\cite{neverova2019correlated}, multi-scale strategy~\cite{guo2019adaptive}, knowledge-transfer~\cite{wang2020ktn}, simulated data~\cite{zhu2020simpose}, or continuous embedding~\cite{neverova2020continuous}. 
All these methods are built upon Mask R-CNN~\cite{he2017mask} framework and follow the two-stage top-down detect-then-segment pipeline. 
In particular, these top-down methods first employ an object detector based on Faster R-CNN~\cite{RenHG017} to predict a bounding box for each person, and then crop and resize regions-of-interest (ROIs) feature maps into a fixed size for each person.
Such a top-down pipeline has several limitations.
(1) The bounding box causes early commitment, \ie if the bounding box fails to detect the whole body, there is no path to recovery, as shown in Fig.~\ref{fig:teaser}. Furthermore, several people may exist in the same bounding box which may not well align with a human due to overlap ambiguity (also mentioned in~\cite{wang2020ktn}), 
(2) The detected person patches are cropped and resized, thus the resolution per person is usually heavily compressed.
(3) Top-down methods cannot fully leverage the sharing computation mechanism of convolutional networks, and thus, their inference time 
scales unfavorably with the number of instances
(see Fig.~\ref{fig:infer_time}).

In contrast, we propose an end-to-end direct method inspired by the success of direct instance segmentation methods~\cite{tian2020conditional,wang2020solo}, which is faster and more accurate under multi-person occlusion than previous methods.
In particular, we formulate the dense pose estimation task into two inter-related sub-tasks: global dense pose IUV  representation
\footnote{The IUV representation, introduced by~\cite{alp2018densepose}, is an image-based UV map with multiple channels.}
estimation and instance segmentation. 
Inferring IUVs in a direct approach avoids early commitment, overlapping ambiguity, and heavy resolution compression, as well as makes computational complexity less dependent on the instance number.
Moreover, a simple fully convolutional implementation (as experimented in~\cite{alp2018densepose}) produces inferior results due to the interference from multi-person feature normalization 
and wastes much computation and GPU memory in the empty background area. To address this, we propose an instance-aware normalization (IAN) technique to improve the results and sparse residual FCN to save computation.
Furthermore, previous methods are designed for per-frame prediction and produce flicker on video sequences. To alleviate flickering, we propose a simple and effective 2D temporal smoothing scheme which naturally fits our direct method. 

We make three contributions:
1) We propose a direct framework DDP for human dense pose estimation, which runs faster and detects humans better in multiple person overlap situations than top-down methods.
2) We propose an instance-aware normalization (IAN) technique to remove the interference from multi-person feature normalization and utilize sparse convolution to skip the background computation.
3) We introduce an effective 2D temporal-smoothing scheme tailored for dense 2D inference problems, which preserves the coherence of projected 2D shapes.

\section{Related Work}
\label{sec:related}

\myparagraph{Instance segmentation.}
Instance segmentation methods combine instance-level object detection and pixel-level semantic segmentation. 
Most existing approaches can be categorized into top-down methods and bottom-up methods. Top-down methods~\cite{li2017fully,he2017mask,liu2018path,bolya2019yolact,huang2019mask,chen2019tensormask} operate in a detect-then-segment way, \ie detect the bounding box and then segment the object within the box. Among these methods, Mask R-CNN~\cite{he2017mask} is the most widely known top-down framework and has seen improved variants~\cite{liu2018path,huang2019mask}. 
Due to the high performance of Mask R-CNN, all prior dense pose methods build upon it.
Bottom-up methods~\cite{newell2017associative,de2017semantic,liu2017sgn} operate in a label-then-cluster way, \eg learn pixel-level embeddings and then cluster them into groups.
Recently, some effective direct methods~\cite{tian2020conditional,wang2020solo,wang2020solov2} 
solved instance segmentation in one stage, \ie without a detect-then-segment or label-then-cluster strategy. 
In their SOLO approach,  Wang~\etal~\cite{wang2020solo} segment objects by locations, \ie using instance location categorization to predict object center locations. 
They further improved accuracy and speed in their improved SOLOv2 method~\cite{wang2020solov2}. 
Tian~\etal propose CondInst~\cite{tian2020conditional}, which employs dynamic instance-aware networks to infer an instance mask from global feature maps directly.
In this work, we take inspiration from CondInst~\cite{tian2020conditional} and propose a new direct framework for dense pose estimation.

\myparagraph{Multi-person 2D pose estimation.}
Multi-person 2D pose estimation methods estimate the number of people, their positions, and their body keypoints (joints) from an image. Similar to instance segmentation methods, multi-person 2D pose estimation algorithms can be classified into top-down and bottom-up methods. 
Top-down methods~\cite{papandreou2017towards,xiao2018simple,huang2017coarse,sun2019deep,wang2020graph,huang2020devil,zhang2020distribution} employ off-the-shelf person detectors to obtain a bounding box for each person and then apply a single-person pose estimator within each box. 
Bottom-up methods~\cite{cao2017realtime,insafutdinov2016deepercut,insafutdinov2017arttrack,newell2017associative,kreiss2019pifpaf,jin2020differentiable,cheng2020higherhrnet} first locate all the body joints in one image, and group them into individual person instances during post-processing.
Both top-down and bottom-up methods require multiple steps to obtain the final keypoint detection results. 

Recently, Tian~\etal propose DirectPose~\cite{tian2019directpose} to handle the multi-person pose estimation. They extend the anchor-free single-stage object detector FCOS~\cite{tian2019fcos}, with one new output branch for keypoint detection. 
Our method DDP follows a similar design principle to avoid the limitations of top-down methods.
Note that, unlike sparse keypoints pose estimation, it is very challenging to formulate the dense correspondence prediction problem as a keypoint classification problem like~\cite{tian2019directpose}.
Thus, we predict the instance mask and global IUV separately. 

\myparagraph{Dense human pose estimation.}
Some methods~\cite{bogo2016keep,kanazawa2018end} estimate pixel-wise correspondence to a 3D surface model by fitting a prior deformable surface model to the image via indirect supervision, \eg through body silhouette and key-joints.
In~\cite{alp2018densepose}, G{\"u}ler~\etal propose to directly map each pixel of a person region in the RGB image to the 3D human surface. 
The dense correspondence is represented in a chart-based format called IUV, \ie 24 body surfaces plus one background category (25 classes in total). With each surface patch, the local correspondence is further represented by a local UV coordinate system. Therefore, their dense pose IUV representation has 25$\times$3=75 dimensions, which are then summarized into three dimensions according to the 25 classes. 
\cite{alp2018densepose} introduces a two-stage DensePose-RCNN built upon Mask R-CNN~\cite{he2017mask} and collects a large-scale image-to-surface sparse mapping dataset for direct supervision. DensePose-RCNN shows promising results on in-the-wild data.
Following this pipeline,  Guo~\etal~\cite{guo2019adaptive} propose a multi-scale method called AMANet with improved performance under scale variations. 
Wang~\etal~\cite{wang2020ktn} improve dense pose estimation by utilizing external commonsense knowledge. 
Neverova~\etal~\cite{neverova2019slim} propose to augment the data annotations with motion cues for performance improvement.
Zhu~\etal~\cite{zhu2020simpose} introduce a new synthetic dense human pose dataset, together with a new estimator using a domain adaptation strategy to achieve good performance on real-world data. 
Recently, Neverova~\etal~\cite{neverova2020continuous} propose a more straightforward and universal representation, Continuous Surface Embeddings (CSE), to better represent dense correspondences and improve the performance.
These works all follow the two-stage top-down strategy that suffers from early commitment, overlap ambiguities, heavy resolution compression, and unfavorable runtime scaling with the number of instances.
In contrast, we propose a direct dense pose estimation framework with faster runtime and better resilience to the aforementioned challenges.

\section{Framework}
\label{sec:framework}

\begin{figure*}[t]
\scriptsize
  \centering
  \includegraphics[width=1\linewidth]{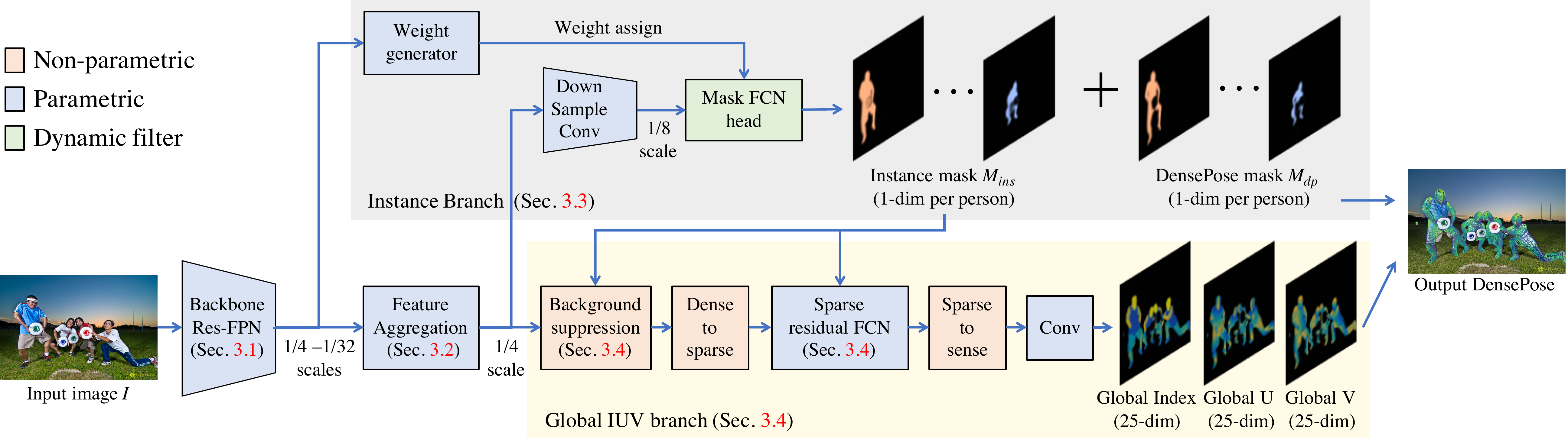}\\
\caption{
Framework overview.
The ResNet-FPN based backbone is first used to extract a feature pyramid.
Then, the feature aggregation module is applied to aggregate the feature pyramid into a global feature representation. Such a global feature is then fed into the instance branch and global IUV branch, respectively, to estimate the instance-level masks and the global IUV representation. 
Note that, for each instance, the Mask FCN weights are generated dynamically via a weight generator.
}
\label{fig:framework}
\end{figure*}

\begin{figure}[t]
\scriptsize
  \centering
  \includegraphics[width=1\linewidth]{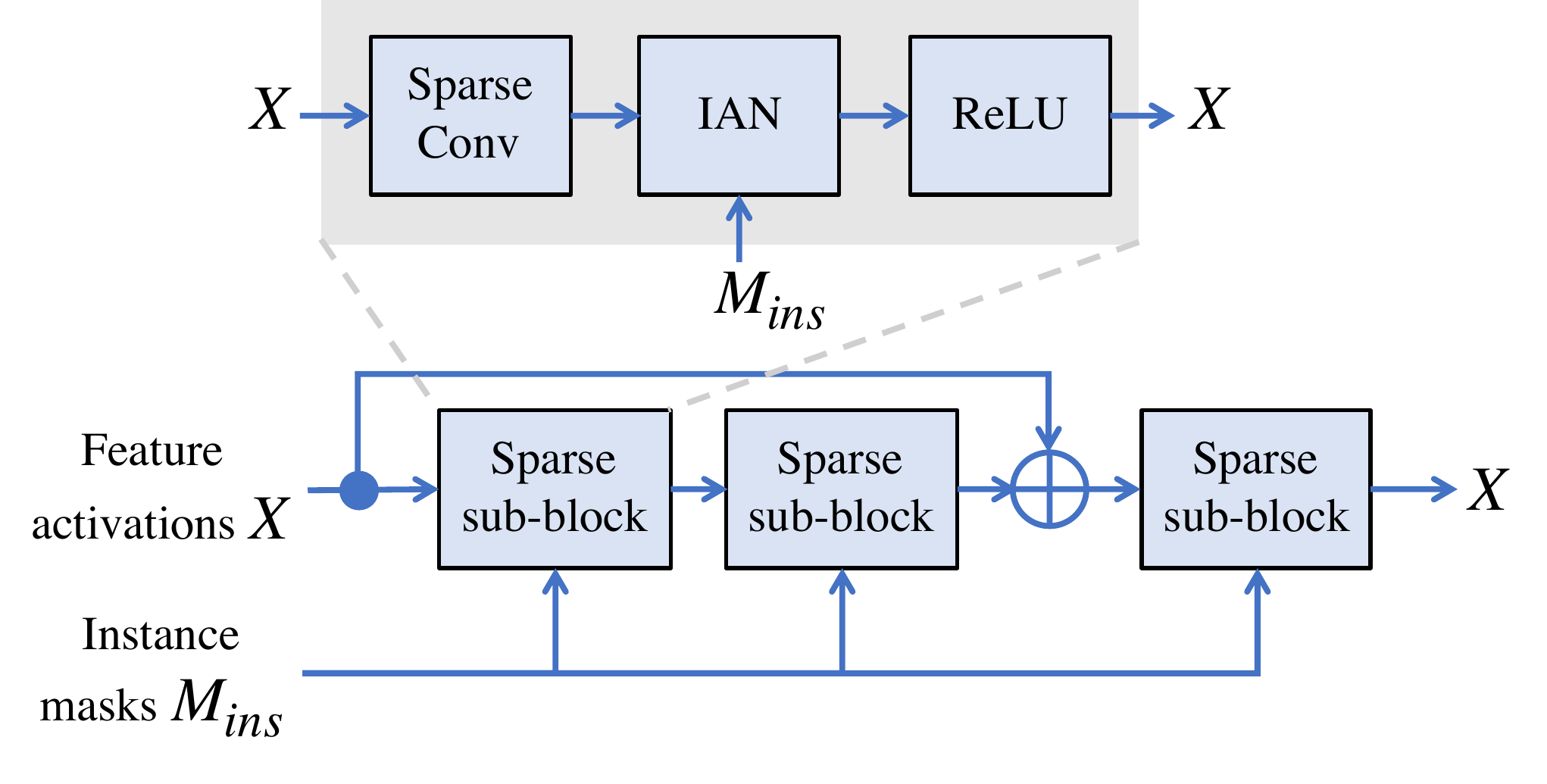}\\
\caption{The sparse residual block contains three sparse sub-blocks. Each sparse sub-block consists of a sparse convolution layer~\cite{graham20183d}, an instance-aware normalization (IAN) layer, and a ReLU layer.}
\label{fig:sparse_block}
\end{figure}

\subsection{Overview}
\label{sec:overview}
We propose a direct framework to divide the multi-person dense pose estimation problem into two subtasks: instance prediction and global IUV dense pose representation prediction.
As illustrated in Fig.~\ref{fig:framework}, our framework starts with a feature extractor backbone - a ResNet network (\eg ResNet-50) followed by a feature pyramid network (FPN)~\cite{lin2017feature}. Then, the extracted feature pyramid (1/4 to 1/32 scales) is aggregated via a feature aggregation (Sec.~\ref{sec:feat_agg}) module, and then fed into the instance branch (Sec.~\ref{sec:instance_branch}) and the global IUV branch (Sec.~\ref{sec:global_iuv_branch}), respectively. The instance branch aims to predict the instance-level information, \ie instance mask $\boldsymbol{M}_{ins}$ and dense pose mask $\boldsymbol{M}_{dp}$, for each person. In
contrast, the global IUV branch aims to predict the dense pose IUV representation for the whole image. The final per-person dense pose estimation can be obtained simply by multiplying the dense pose mask $\boldsymbol{M}_{dp}$ with the global IUV representation in an element-wise way.
In addition, we also propose a 2D temporal-smoothing scheme (Sec.~\ref{sec:smoothing}) to alleviate the temporal jitters when dealing with video data.

\subsection{Feature aggregation module}
\label{sec:feat_agg}
To alleviate instance scale variation issue in our direct method, we adopt a feature pyramid aggregation module~\cite{kirillov2019panoptic} to aggregate features of different scales. Specifically, each FPN level is upsampled by convolutions and bilinear upsampling until it reaches 1/4 scale. These upscaled features are  summed together as $\boldsymbol{X}_{agg}$ which are fed into our instance branch and global IUV branch, respectively.

\subsection{Instance branch}
\label{sec:instance_branch}
The instance branch is built on top of a direct instance segmentation method CondInst~\cite{tian2020conditional} whose core idea is to dynamically generate the specific mask FCN head parameters for each instance\footnote{Similar to CondInst~\cite{tian2020conditional}, there are centerness and box heads in our instance branch, but omitted in the figure for simplicity.}. 
For an image with $K$ instances, $K$ different mask FCN heads will be dynamically generated, each containing the characteristics of its target instance in the filters. 
In particular, as shown in Fig.~\ref{fig:framework} top, 
the instance branch first processes the received aggregated features $\boldsymbol{X}_{agg}$ with a Down Sample Conv module to extract global instance features $\boldsymbol{X}_D$ and reduce the feature resolution to 1/8 scale for saving computation. 
This Down Sample Conv module contains three convolution layers of stride 1 and one convolution layer with stride 2. 
Moreover, $\boldsymbol{X}_D$ is combined with a map of the coordinates, which are relative coordinates from all the locations on $\boldsymbol{X}_D$ to the location (h, w) (\ie where the filters of the mask head are generated). Then, the combination, termed as $\Tilde{\boldsymbol{X}}_D$, is sent to the mask FCN head, whose parameters $\theta_{h,w}$ is dynamically generated by the weight generator module, to predict the instance mask.
In other words, $\Tilde{\boldsymbol{X}}_D$ contains the global information for the whole image, while the parameters $\theta_{h,w}$ encode instance-aware characteristics (\eg relative position, shape, and appearance).
Different from CondInst, our mask FCN head will produce the instance mask $\boldsymbol{M}_{ins}$ and dense pose mask $\boldsymbol{M}_{dp}$ jointly for each instance. 
This joint learning design takes the advantages of the existing rich instance annotations for stabilizing training (explained in Sec.~\ref{sec:learning}), as well as provides instance masks for the Global IUV branch (explained in Sec.~\ref{sec:global_iuv_branch}).
Note that one can also adopt other advanced direct instance segmentation networks for our instance branch, like SOLO~\cite{wang2020solo} and SOLOv2~\cite{wang2020solov2}.

\subsection{Global IUV branch}
\label{sec:global_iuv_branch}
We aim to predict the dense pose IUV information globally where all IUVs of different people are represented in the same image plane. 
However, merely performing convolution over the large background region is computationally quite expensive and wasteful. Therefore, we propose a background suppression operation followed by a sparse residual FCN module to only deal with the person region of interest. We also  propose an instance-aware normalization (IAN) technique that performs better than normalizing all the instances together as done in~\cite{alp2018densepose}.

\myparagraph{Background suppression.}
As for the human dense pose estimation task, we are only interested in the people region, which usually occupies a small part of the whole image. For example, the sparsity (\ie ratio of the people region over the whole image) on the DensePose-COCO dataset~\cite{alp2018densepose} is on average  23.7\% and ranges from 1.7\% to 90.9\%\footnote{Calculated on the instance annotations of DensePose-COCO minival split.}. Some sparsity examples are illustrated in Fig.~\ref{fig:teaser}.
These large background regions will waste a large amount of computation and interfere with the learning procedure as well.
Previous top-down multi-person dense pose estimation methods~\cite{alp2018densepose,guo2019adaptive,neverova2019correlated,wang2020ktn,zhu2020simpose,wu2019detectron2,neverova2020continuous} alleviate this issue by cropping the feature maps according to the detected bounding boxes and process them separately in a single person style.
Such a cropping-based approach is sub-optimal as it still introduces some background or content from other instances. Besides, this suffers from the early commitment issue.
In contrast, we propose a background suppression strategy to suppress the background interference explicitly. 
Specifically, we first resize the estimated instance masks $\{\boldsymbol{M}_{ins}^i\}_{i=1,\dots,N}$ to the resolution of features $\boldsymbol{X}_{agg}$. The resized masks $\{\Tilde{\boldsymbol{M}}_{ins}^i\}_{i=1,\dots,N}$ are then combined into a foreground mask $\boldsymbol{M}_{fg}$ which is applied to mask the features in a point-wise manner as follows,
\begin{align}
    \boldsymbol{X}_{fg}=\boldsymbol{M}_{fg} \odot \boldsymbol{X}_{agg}, 
    ~\text{where}~ \boldsymbol{M}_{fg}=\bigcup\limits_{i=1}^{N} \Tilde{\boldsymbol{M}}_{ins}^i,
\end{align}
where $\boldsymbol{X}_{fg}$ denote the masked foreground features.

\myparagraph{Sparse residual FCN.}
As mentioned above, computing the features on background regions is wasteful, especially considering that we need to maintain high-resolution features to achieve dense global IUV prediction. Therefore, we perform convolution in a sparse manner to reduce the computation and memory cost. Specifically, the dense features are first transformed into a sparse format and then processed by our sparse residual FCN module and finally transformed back to a dense format.
The sparse residual FCN module consists of three sparse residual blocks. As illustrated in Fig.~\ref{fig:sparse_block}, each sparse residual block contains two sparse sub-blocks with a residual skip connection followed by a third sparse sub-block (except for the last sparse residual block). Each sparse sub-block includes a sparse convolution layer, an instance-aware normalization (IAN) layer, and a ReLU layer.
Regarding the sparse convolution layer, we adopt the submanifold sparse convolution (SSC) layer~\cite{graham20183d}, which can fix the location of active sites and thus maintain the same level of sparsity throughout the network. 
The instance-aware normalization (IAN) layer is described in the next paragraph.

\myparagraph{Instance-aware normalization.}
We propose the instance-aware normalization to perform feature normalization for each instance separately.
In particular, we integrate the resized instance masks $\{\Tilde{\boldsymbol{M}}_{ins}\}_{i=1,\dots,N}$ into our normalization operation as follows,
\begin{align}
    \mu_{nci} &= \frac{1}{\text{Numel}(\Tilde{\boldsymbol{M}}_{ins}^i)} \sum\limits_{h,w \in \Tilde{\boldsymbol{M}}_{ins}^i} x_{nchw}, 
    \label{eq:IAN_mu}  \\
    \sigma^2_{nci} &= \frac{1}{\text{Numel}(\Tilde{\boldsymbol{M}}_{ins}^i)} \sum\limits_{h,w \in \Tilde{\boldsymbol{M}}_{ins}^i} (x_{nchw}-\mu_{nci})^2, 
    \label{eq:IAN_sigma}  \\
    \hat{x} &= \bigcup\limits_{i=1}^{N} \bigcup\limits_{h,w \in \Tilde{\boldsymbol{M}}_{ins}^i} \Big( \frac{x_{nchw}-\mu_{nci}}{\sqrt{\sigma^2_{nci}+\epsilon}} \gamma + \beta \Big),
    \label{eq:IAN}
\end{align}
where $x_{nchw}$ denotes the feature point at location (h,w) exists in $\Tilde{\boldsymbol{M}}_{ins}^i$, \ie the region of the $i$-th instance. Numel($\cdot$) means the point number in the region of the $i$-th instance.
We calculate the mean $\mu_{nci}$ and variance $\sigma^2_{nci}$ for each instance mask $\Tilde{\boldsymbol{M}}_{ins}^i$ individually, and then apply instance normalization~\cite{ulyanov2016instance} due to its better performance over other normalization techniques in our pilot experiment. Note that, all the instances share the same learnable affine parameters $\gamma$ and $\beta$ as they all belong to the person category.


\section{2D temporal-smoothing scheme}
\label{sec:smoothing}
Previous dense pose estimation methods are designed for image-based prediction because it is expensive to collect accurate dense pose annotations for video data. However, directly applying image-based methods to video data leads to the undesirable flickering issue, since the temporal information is not considered. To address this, we introduce a simple and effective 2D temporal-smoothing scheme, which fits well our global IUV representation.
The main idea is to use the temporal constraint from the original RGB video. 
Specifically, given the present RGB frame $I_t$ and its temporally adjacent frames $\{{I_{t+j}}\}_{j=-r,\cdots,-1,1,\cdots,r}$,
we adopt an optical flow estimation model (\eg RAFT~\cite{teed2020raft}) to predict the optical flows $\{f_{t\rightarrow t+j}\}_{j=-r,\cdots,-1,1,\cdots,r}$, which are  used to warp the global IUV dense pose representation $\boldsymbol{C}$  of adjacent frames to one $\boldsymbol{C}_{temp}$ for the present frame using the following weighted sum:
\begin{align}
    \boldsymbol{C}_{t+j \rightarrow t} &= \text{Warp}(\boldsymbol{C_{t+j}}, f_{t\rightarrow t+j}),
    \label{eq:warp} \\
    \boldsymbol{C}_{temp} &= \sum\limits_{j=-r}^r \alpha_j \boldsymbol{C}_{t+j \rightarrow t},
    \label{eq:IUV_temp}
\end{align}
where $r$ is the temporal window interval and $\alpha_j$ is the sum weight. We set $r=2$ and $\{\alpha_j\} = [0.2,0.2,0.2,0.2,0.2]$ by default. 
Here we perform the warping and weighted sum operations on the continuous logit-level, \ie $\boldsymbol{C}$ is a 75-dim logit representation.


\section{Learning}
\label{sec:learning}
As mentioned before, we predict instance mask $\boldsymbol{M}_{ins}$ and dense pose mask $\boldsymbol{M}_{dp}$ jointly in our instance branch (Sec.~\ref{sec:instance_branch}). 
We take this joint approach because of the limited dense pose annotation and the limited body coverage. 
Regarding the annotation, only part of human instances was selected for dense pose annotation on the DensePose-COCO dataset. Training with  annotated dense pose masks only will lead to inferior instance prediction performance. 
Furthermore, the dense pose masks do not cover the whole person, leading to important information missing in our background suppression. (Sec.~\ref{sec:global_iuv_branch}).
Therefore, we jointly regress both instance masks and dense pose masks in our instance branch to improve the learning with rich instance annotations and suppress the background with instance masks.
Besides, we apply the ground truth instance masks for background suppression in our global IUV branch during training, considering that the estimated instance masks contain many errors in the early training stage. Such errors will break the learning of the global IUV branch.

\begin{table*}[t]
    \centering 
    \setlength{\tabcolsep}{4.3pt} 
    \begin{tabular}
    {@{\extracolsep{\fill}} l| c c c| c c| c c c| c c|c}
    \toprule 
    Method & AP & AP$_{50}$ & AP$_{75}$ & AP$_{M}$ & AP$_{L}$ & AR & AR$_{50}$ & AR$_{75}$ & AR$_{M}$ & AR$_{L}$ & time (s) \\
    \midrule[0.6pt]	
    \multicolumn{11}{c}{Top-down methods} \\
    \midrule[0.6pt]	
    DP-cascade (Res50)~\cite{alp2018densepose} & 51.6 & 83.9 & 55.2 & 41.9 & 53.4 & 60.4 & 88.9 & 65.3 & 43.3 & 61.6 & 0.583 \\
    DP-cascade (Res50)+masks~\cite{alp2018densepose} & 52.8 & 85.5 & 56.1 & 40.3 & 54.6 & 62.0 & 89.7 & 67.0 & 42.4 & 63.3 & - \\
    DP-cascade (Res50)+keypoints~\cite{alp2018densepose} & 55.8 & 87.5 & 61.2 & 48.4 & 57.1 & 63.9 & 91.0 & 69.7 & 50.3 & 64.8 & - \\
    Parsing (Res50)~\cite{yang2019parsing} & 55.0 & 87.6 & 59.8 & 50.6 & 56.6 & - & - & - & - & - & 0.392 \\
    Parsing (Res50)+keypoints~\cite{yang2019parsing} & 58.3 & 90.1 & 66.9 & 51.8 & 61.9 & - & - & - & - & - & - \\
    Parsing (ResNeXt101)~\cite{yang2019parsing} & 59.1 & 91.0 & 66.9 & 51.8 & 61.9 & - & - & - & - & - & - \\
    Parsing (ResNeXt101)+keypoints~\cite{yang2019parsing} & 61.6 & 91.6 & 72.3 & 54.8 & 64.8 & - & - & - & - & - & - \\
    SimPose (Res50)*~\cite{zhu2020simpose} & 57.3 & 88.4 & 67.3 & 60.1 & 59.3 & 66.4 & 95.1 & 77.8 & 62.4 & 66.7 & - \\
    AMA-Net (Res50)~\cite{guo2019adaptive} & 64.1 & 91.4 & 72.9 & 59.3 & 65.3 & 71.6 & 94.7 & 79.8 & 61.3 & 72.3 & - \\
    KTN (Res50)~\cite{wang2020ktn} & 66.5 & 91.5 & 75.5 & 61.9 & 68.0 & 74.2 & 95.2 & 82.3 & \textbf{64.2} & \textbf{74.9} & 0.518 \\
    DP R-CNN DeepLab (Res50)~\cite{neverova2020continuous} & 66.8 & 92.8 & \textbf{79.7} & 60.7 & 68.0 & 72.1 & 95.8 & 82.9 & 62.2 & 72.4 & 0.242 \\
    DP R-CNN DeepLab (Res101)~\cite{neverova2020continuous} & \textbf{67.7} & \textbf{93.5} & \textbf{79.7} & \textbf{62.6} & \textbf{69.1} & \textbf{73.6} & \textbf{96.5} & \textbf{84.7} & \textbf{64.2} & 74.2 & 0.382 \\
    \midrule[0.6pt]	
    \multicolumn{11}{c}{Direct methods} \\
    \midrule[0.6pt]	
    Ours (Res50) & 64.0 & 92.4 & 76.0 & 57.2 & 65.7 & 70.9 & 96.4 & 82.4 & 59.9 & 71.6 & \textbf{0.209} \\
    \bottomrule[1pt]
    \end{tabular}
    \caption{The quantitative results on DensePose-COCO minival split. * Models are trained with a simulated dataset.
    }
    \label{tab:comparisons}
\end{table*}

\myparagraph{Losses.}
The overall loss function of DDP is formulated as,
\begin{align}
    L_{all} &= L_{mask} + L_{IUV},
    \label{eq:loss_all} \\
    L_{mask} &= L_{fcos} + \lambda_1 (L_{M_{ins}} + L_{M_{dp}})
    \label{eq:loss_ins} \\
    L_{IUV} &= L_I + \lambda_2 L_{UV} + \lambda_3 L_s
    \label{eq:loss_IUV} 
\end{align}
where $L_{mask}$ and $L_{IUV}$ denote the loss for instance branch and global IUV branch, respectively. 
$\lambda_1=5$, $\lambda_2=10$, $\lambda_3=1$ are used to balance the losses.
Similar to CondInst, 
our instance prediction loss $L_{mask}$ includes $L_{fcos}$ and two mask losses $L_{M_{ins}}$, $L_{M_{dp}}$.
We refer the reader to FCOS~\cite{tian2019fcos} for the details of $L_{fcos}$. 
$L_{M_{ins}}$ and $L_{M_{dp}}$ are defined as,
%
\begin{align}
    \resizebox{1\hsize}{!}{$
        L_{M_{ins}} = \frac{1}{N_{pos}} \sum\limits_{h,w} \mathbbm{1}_{\{c^*_{h,w}>0\}} L_{dice} (\boldsymbol{M}^{h,w}_{ins},\boldsymbol{M}^{h,w*}_{ins}), 
    $}
    \label{eq:mask_ins} \\
    \resizebox{1\hsize}{!}{$
        L_{M_{dp}} = \frac{1}{N_{pos}} \sum\limits_{h,w} \mathbbm{1}_{\{c^*_{h,w}>0\}} L_{dice} (\boldsymbol{M}^{h,w}_{dp},\boldsymbol{M}^{h,w*}_{dp}), 
    $}
    \label{eq:mask_dp}
\end{align}
where $[\boldsymbol{M}^{h,w}_{ins},\boldsymbol{M}^{h,w}_{dp}]=\text{MaskHead}(\Tilde{\boldsymbol{X}}_D;\theta_{h,w})$ are the estimated instance and dense pose masks, and $\boldsymbol{M}^{h,w*}_{ins},\boldsymbol{M}^{h,w*}_{dp}$ are the corresponding ground truth.
The $c^*_{h,w}$ is the classification label of location (h, w), which is the class of the instance associated with the location or 0 (\ie background) if the location is not
associated with any instance. $N_{pos}$ is the number of locations where $c^*_{h,w}>0$.
Our global dense pose IUV prediction loss $L_{IUV}$ contains one cross-entropy loss $L_I$ for body parts classification, one smooth L1-loss $L_{UV}$ for local UV coordinates regression, and one smoothing loss $L_s$. 
The smoothing loss $L_s$ is used to encourage the model to produce less noisy IUV representation, since the dense pose annotation is a set of sparse points. In particular, we adopt an edge-aware smoothness regularization~\cite{godard2017unsupervised} as follows,
\begin{align}
    L_s = \frac{1}{N} \sum\limits_{i=1}^N &|\nabla_h \boldsymbol{C}|e^{-|\nabla_h \boldsymbol{M}_{ins}^i|} \nonumber \\
    + &|\nabla_w \boldsymbol{C}|e^{-|\nabla_w \boldsymbol{M}_{ins}^i|},
    \label{eq:loss_smooth}
\end{align}
where $\boldsymbol{C}$ is the predicted global IUV representation.

\section{Experiments}
\label{sec:exp}

For evaluation, we present qualitative and quantitative results, a runtime analysis, and user study results.

\begin{table*}[t]
    \centering 
    \setlength{\tabcolsep}{3.7pt} 
    \begin{tabular}
    {@{\extracolsep{\fill}} l| c c c| c c| c c c| c c | c}
    \toprule 
    Method & AP & AP$_{50}$ & AP$_{75}$ & AP$_{M}$ & AP$_{L}$ & AR & AR$_{50}$ & AR$_{75}$ & AR$_{M}$ & AR$_{L}$ & time (s) \\
    \midrule[0.6pt]	
    Ours (Res50) w/o $L_s$, w/o IAN, w/o Sparse & 62.2 & 91.5 & 73.4 & 55.9 & 64.0 & 69.3 & 95.9 & 80.3 & 58.2 & 70.1 & 0.380 \\ 
    Ours (Res50) w/o $L_s$, w/o IAN & 62.2 & 92.3 & 73.2 & 55.0 & 63.9 & 69.3 & 96.2 & 80.4 & 58.9 & 70.0 & 0.234 \\
    Ours (Res50) w/o $L_s$ & 63.8 & 91.8 & 75.9 & 57.1 & \textbf{65.8} & \textbf{71.2} & 96.2 & \textbf{83.4} & 59.0 & \textbf{71.9} & 0.211 \\
    \midrule[0.6pt]	
    Ours (Res50) & \textbf{64.0} & \textbf{92.4} & \textbf{76.0} & \textbf{57.2} & 65.7 & 70.9 & \textbf{96.4} & 82.4 & \textbf{59.9} & 71.6 & \textbf{0.209} \\ 
    \bottomrule[1pt]
    \end{tabular}
    \caption{Ablation study and average inference time on DensePose-COCO minival split.
    }
    \label{tab:ablation}
\end{table*}

\begin{table}[t]
    \centering 
    \setlength{\tabcolsep}{5.5pt} 
    \begin{tabular}
    {@{\extracolsep{\fill}} l c}
    \toprule 
    Method & Human preference \\
    \midrule[0.6pt]	
    DP R-CNN DeepLab (Res50)~\cite{neverova2020continuous} & 0.076 \\
    Ours (Res50) w/ temporal smoothing & 0.924 \\ 
    \midrule[0.6pt]	
    Ours (Res50) w/o temporal smoothing & 0.016 \\ 
    Ours (Res50) w/ temporal smoothing & 0.984 \\ 
    \bottomrule[1pt]
    \end{tabular}
    \caption{User study for temporal smoothing.
    }
    \label{tab:user_study}
\end{table}

\myparagraph{Implementation details.}
Unless specified otherwise, we use the following implementation details.
We use a ResNet-50~\cite{he2016deep} architecture as the backbone, followed by a 4-level FPN (\ie 1/4, 1/8, 1/16, 1/32 levels). The ResNet weights are initialized by the pre-trained keypoints estimation models from COCO~\cite{lin2014microsoft}. Our models are trained with stochastic gradient descent
(SGD) for 130K iterations with an initial learning rate of 0.01 and a mini-batch of 8 images. The learning rate is reduced by a factor of 10 at iteration 100K and 120K, respectively. Weight decay and momentum are set as 0.0001 and 0.9, respectively. 
Following~\cite{alp2018densepose}, two quantitative metrics are used for evaluation, \ie
Average Precision (AP) and Average Recall (AR). Both metrics are calculated at a number of geodesic point similarity (GPS) ranging from 0.5 to 0.95. In addition, other evaluation metrics, \ie AP$_M$ and AR$_M$ for medium people and AP$_L$ and AR$_L$ for large people, are also reported.
Our method is implemented based on the Detectron2 framework~\cite{wu2019detectron2}. 
%
No data augmentation is used during training or testing. 
Inference times are measurd on a single V100 GPU with one image per batch.

\myparagraph{Dataset.}
Our method is evaluated on DensePose-COCO dataset~\cite{alp2018densepose}, 
which has manually annotated correspondences on a subset of the COCO dataset~\cite{lin2014microsoft}. 
There are about 50K labeled human instances 
each of which is annotated with 100 points on average. In total, there are about 5 million manually annotated correspondences. The dataset is split into a training set and a validation set with 32K images and 1.5k images, respectively.

\subsection{Comparisons}
We evaluate the proposed end-to-end DDP framework on DensePose-COCO minival split and compare it with the SOTA top-down methods in Table~\ref{tab:comparisons}. Our method achieves 64.0\% AP and 70.9\% AR with ResNet-50.
The performance of our model is better than the strong baselines DP-cascade~\cite{alp2018densepose} (64.0\% vs. 55.8\% AP) and Parsing~\cite{yang2019parsing} (64.0\% vs. 61.6\% AP), and is comparable to AMA-Net~\cite{guo2019adaptive} (64.0\% vs. 64.1\% AP).
Our model is still behind the top-performing top-down methods KTN~\cite{wang2020ktn} and DP R-CNN DeepLab~\cite{neverova2020continuous}. 
Nevertheless, it is worth noting that the top-down strategy suffers from the early commitment issue, overlap ambiguities as shown in Fig.~\ref{fig:teaser}.
Besides, top-down methods run slower in larger multi-person scenes and compress image resolution heavily.

\begin{figure}[t]
\scriptsize
  \centering
  \includegraphics[width=0.8\linewidth]{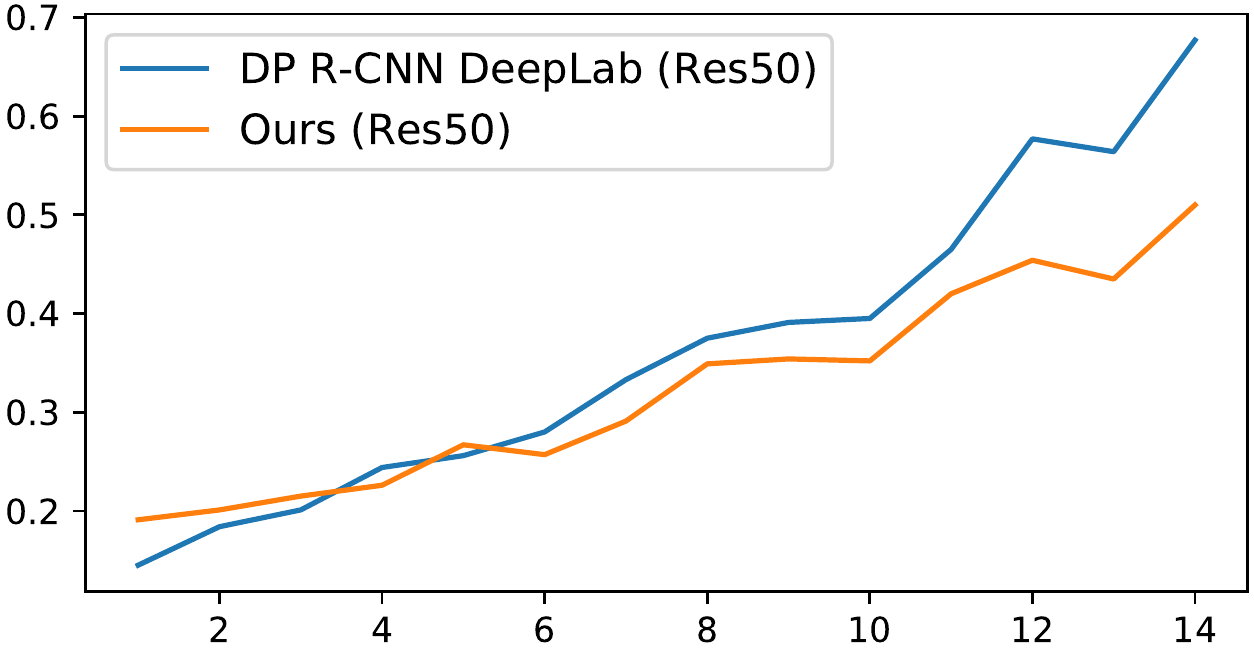}
\caption{Inference time. X-axis: number of person instances on each image. Y-axis: average inference time in seconds for each image.
}
\label{fig:infer_time}
\end{figure}

\begin{figure}[t]
  \centering
  \includegraphics[width=1\linewidth]{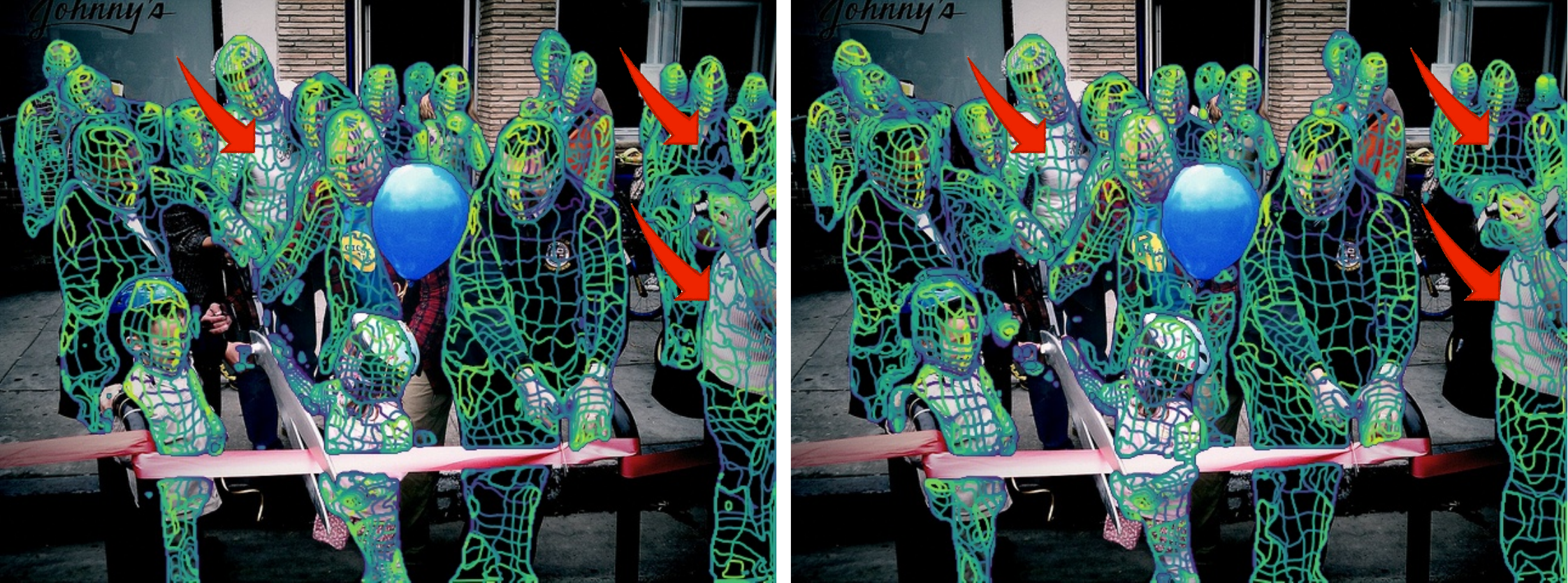}\\ 
  Ours (Res50) w/o $L_s$ ~~~~~~~~~~~~~~~~~~ Ours (Res50)~~~~~
\caption{Effectiveness of smoothing loss $L_s$.
}
\label{fig:smooth}
\end{figure}

\subsection{Ablation study and inference time}
We perform an ablation study for each component of DDP presented in Table~\ref{tab:ablation}. The ablative settings are as follows. \textbf{w/o $L_{smooth}$}: Removing the smoothing loss. \textbf{w/o IAN}: Removing the instance-aware normalization. \textbf{w/o Sparse}: We use full dense feature maps instead of sparse feature maps, \ie $\boldsymbol{M_{fg}}$ is a tensor with all elements set to one.
We observe that the proposed IAN technique improves the results, and the sparse technique aids to significantly reduce the computation time.
Regarding the smoothing loss $L_s$, the quantitative results are almost the same, while the qualitative results are spatially smoothed. As shown in the isocontour visualization in Fig.~\ref{fig:smooth}, the isocontour becomes less noisy when applying the smoothing loss $L_s$.
Such smoothness is desirable because of the continuity nature of human body surface.
Furthermore, we also compare the inference time on images of different instance numbers with the state-of-the-art top-down method DP R-CNN DeepLab~\cite{neverova2020continuous} as shown in Fig.~\ref{fig:infer_time}.
We observe that our method's inference time increases more slowly with the number of people in the scene,
because our approach's inference time mainly depends on the image's sparsity. 

\begin{figure*}[t]
\scriptsize
  \centering
  \includegraphics[width=1\linewidth]{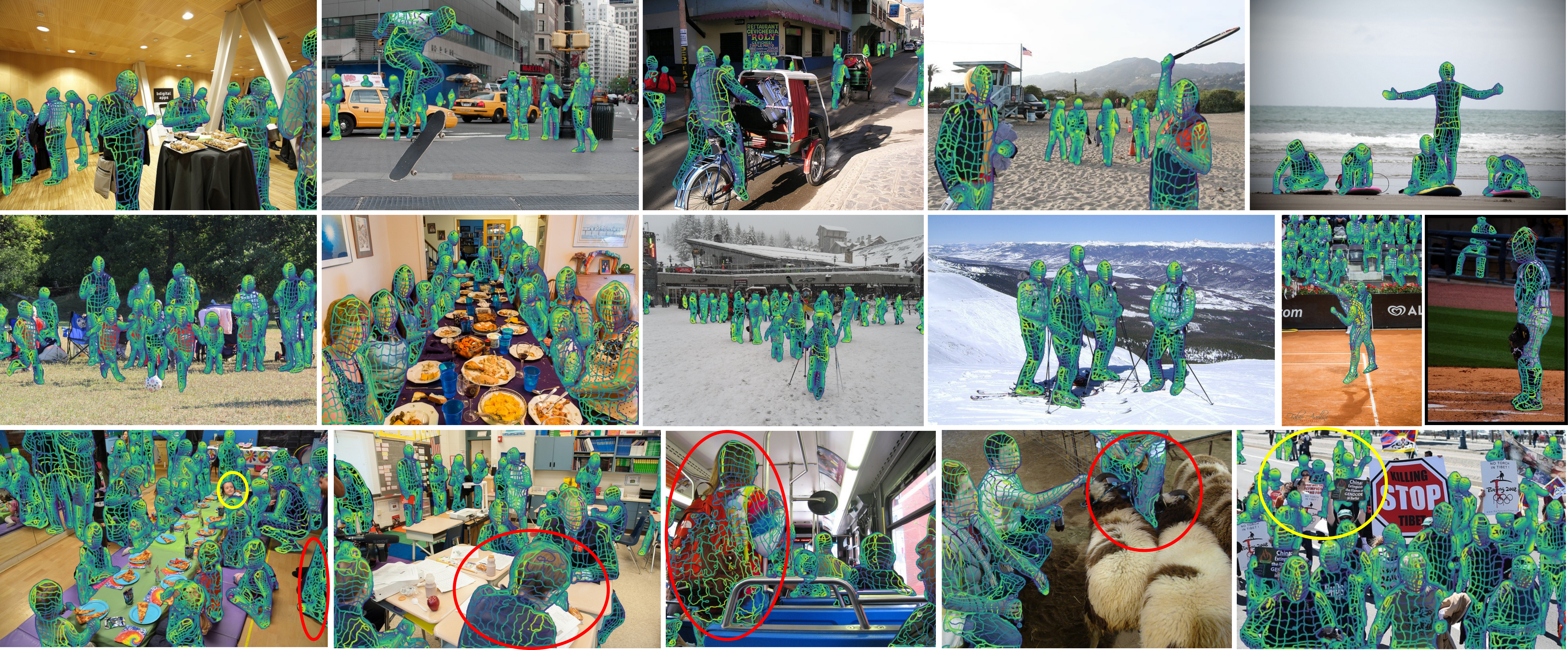}\\
\caption{Qualitative results on DensePose-COCO minival split. The red and yellow circles spot the failure cases (see Sec.~\ref{sec:limitation}).}
\label{fig:qualitative}
\end{figure*}

\subsection{Temporal smoothing evaluation}
\myparagraph{User study.} To demonstrate the effectiveness of our 2D temporal-smoothing scheme, we perform a user study on 11 YouTube video sequences of 20 seconds each. 
As shown in Table~\ref{tab:user_study}, we ask 30 users to assess the temporal smoothing of the results of two comparisons: (1) our method versus the top-performing top-down method - DP R-CNN DeepLab (Res50)~\cite{neverova2020continuous} (row 1-2), and (2) our method versus our non-smoothing setting (row 3-4). 
For each video sequence, given two randomly ordered video results of dense pose isocontour visualization from two methods, users are asked to pick one that looks temporally smoother than the other. The results in Table~\ref{tab:user_study} illustrate that our method is obviously preferred over the alternative.

\myparagraph{Quantitative evaluation.}
Furthermore, we also perform a quantitative evaluation measured with the most widely used metrics in the video stabilization field: Interframe Transformation Fidelity (ITF) index~\cite{marcenaro2001image} and Interframe Similarity Index (ISI)~\cite{guilluy2018performance}, which are based on the video inter-frame PSNR and SSIM scores, respectively. 
We report the average scores (see Tab.~\ref{tab:temporal_eval}) of 11 isocontour visualization videos used in the user study. We can see that our temporal smoothing strategy improves the temporal smoothness on both evaluation metrics.

\myparagraph{Image-to-image translation.} We also apply our method on the human pose transfer task,
\ie translate the dense pose IUV representation input to an RGB image.
We adopt the popular pix2pixHD~\cite{wang2018high} as our translation model and generate the image frame-by-frame.
We observe that our temporally smooth dense pose can help alleviate video flickering issue and generate stable video results.
Please refer to our supplementary video for the visualization results.

\begin{table}[t]
    \centering 
    \setlength{\tabcolsep}{5.5pt} 
    \begin{tabular}
    {@{\extracolsep{\fill}} l c c}
    \toprule 
Model & ITF & ISI \\
\midrule[0.6pt]	
	DP R-CNN DeepLab (Res50)~\cite{neverova2020continuous} & 35.04 & 0.9202 \\
	Ours (Res50) w/o temporal smoothing & 35.24 & 0.9253 \\
    Ours (Res50) w/ temporal smoothing  & \textbf{35.68} & \textbf{0.9294} \\
\bottomrule[1pt]
    \end{tabular}
    \caption{Quantitative evaluation for temporal smoothing.
    }
    \label{tab:temporal_eval}
\end{table}

\begin{figure}[t]
  \centering
  \scriptsize
  \includegraphics[width=1\linewidth]{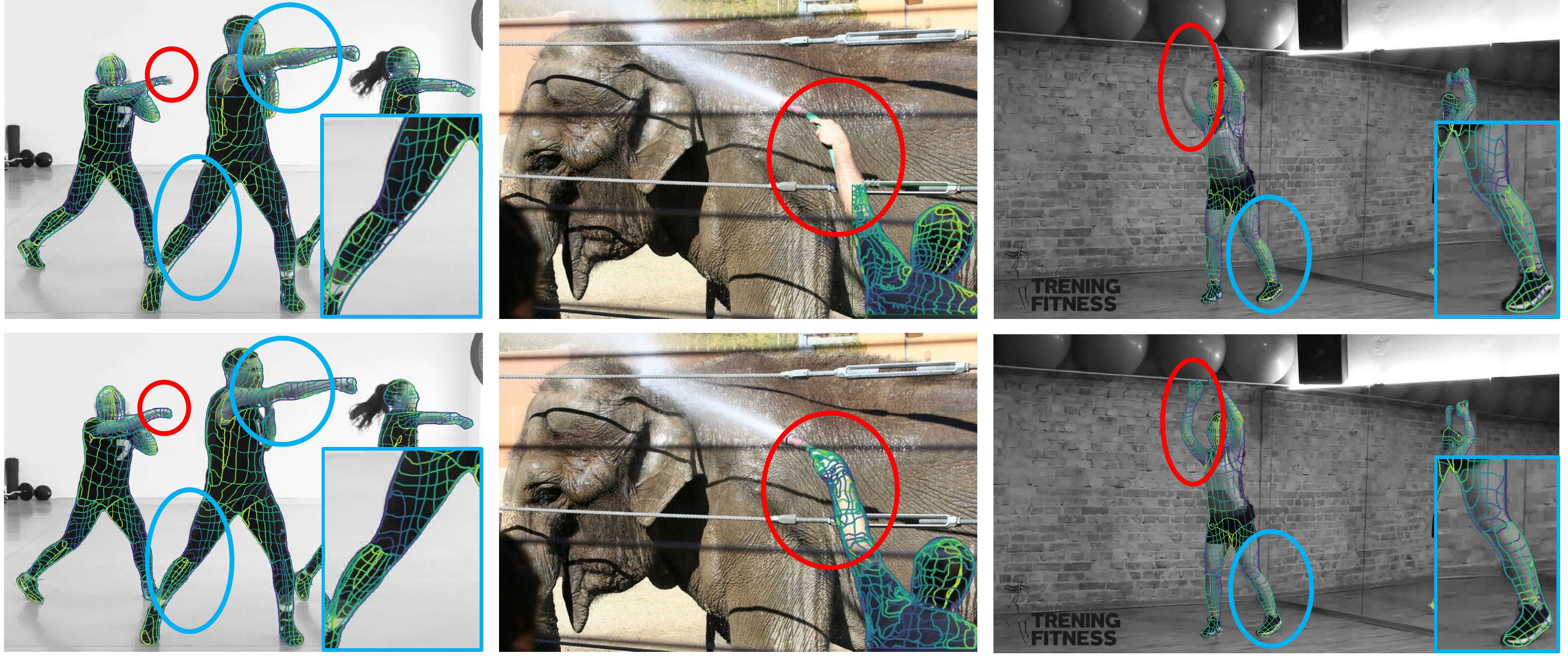}
\caption{Results of~\cite{neverova2020continuous} (top) and Ours (bottom) with Res50. The red and blue circles spot our advantages.
}
\label{fig:compare}
\end{figure}

\section{Qualitative results and discussion}
\label{sec:limitation}
In Fig.~\ref{fig:compare}, we show the advantages of our method compared to the top-down methods~\cite{neverova2020continuous} of avoiding early commitment (red circles) and better boundaries (blue circles).
In Fig.~\ref{fig:qualitative}, we illustrate more isocontour visualization results. We observe that our DDP can predict smooth and accurate 3D body correspondences in diverse real-world scenarios exhibiting different challenges, \eg illumination variations (indoor and outdoor),  occlusions (self occlusion, inter occlusion, and background occlusion), diverse body poses and views. 
Although our method achieves comparable performance to SOTA methods, it may produce noisy results for large-scale or occluded people (\eg the red circle in Fig.~\ref{fig:qualitative}) and fail to detect some occluded small people (\eg the yellow circle in Fig.~\ref{fig:qualitative}). 
As for our 2D temporal-smoothing scheme, it can alleviate the temporal flickering for most cases, but may fail in textureless region or varying illumination where the optical flow estimation is not robust.

\section{Conclusion}
We present a direct human dense pose estimation method DDP that deals with the multi-person dense pose estimation via two inter-related sub-tasks: global IUV estimation and instance segmentation. The proposed method achieves comparable results to strong 
top-down benchmark methods, and avoids issues like early commitment and overlap ambiguity. Our method also runs in weak instance number dependent time. 
Furthermore, we introduce a temporal smoothing pose-processing which naturally fits the global IUV representation and enables temporal smoothing dense pose estimation. We believe such a temporal smoothing ability can benefit human analysis and synthesis tasks.

\noindent\textbf{Acknowledgements.} Liqian Ma and Luc Val Gool were supported by Toyota Motors Europe. Christian Theobalt was supported by ERC Consolidator Grant 4DReply (770784). Lingjie Liu was supported by Lise Meitner Postdoctoral Fellowship.

\small
\bibliographystyle{ieee_fullname}
\bibliography{egbib}

\end{document}


\title{Divide-and-conquer Direct Dense Pose Estimation -- Supplementary Material}

\author{First Author\\
Institution1\\
Institution1 address\\
{\tt\small firstauthor@i1.org}
\and
Second Author\\
Institution2\\
First line of institution2 address\\
{\tt\small secondauthor@i2.org}
}

\maketitle
\ificcvfinal\thispagestyle{empty}\fi

In this supplementary material, we provide additional results, more visual comparisons with prior art and implementation details. 
